\documentclass{article}
\usepackage{spconf,amsmath,graphicx}
\usepackage{makecell,amssymb}
\usepackage{subfigure}

\title{Lip-reading with Hierarchical Pyramidal Convolution and Self-Attention}
\name{Hang Chen$^{\star}$, Jun Du$^{\star}$, Yu Hu$^{\star}$, Li-Rong Dai$^{\star}$, Chin-Hui Lee$^{\dagger}$, Bao-Cai Yin$^{\ddagger}$}
\address{$^{\star}$ National Engineering Laboratory for Speech and Language Information Processing,\\ University of Science and Technology of China \\ $^{\dagger}$ School of Electrical and Computer Engineering, Georgia Institute of Technology\\$^{\ddagger}$iFlytek Research, iFlytek Co., Ltd.}
\begin{document}
%
\maketitle
\begin{abstract}
In this paper, we propose a novel deep learning architecture to improving word-level lip-reading. On the one hand, we first introduce the multi-scale processing into the spatial feature extraction for lip-reading. Specially, we proposed hierarchical pyramidal convolution (HPConv) to replace the standard convolution in original module, leading to improvements over the model’s ability to discover fine-grained lip movements. On the other hand, we merge information in all time steps of the sequence by utilizing self-attention, to make the model pay more attention to the relevant frames. These two advantages are combined together to further enhance the model’s classification power. Experiments on the Lip Reading in the Wild (LRW) dataset show that our proposed model has achieved $86.83\%$ accuracy, yielding $1.53\%$ absolute improvement over the current state-of-the-art. We also conducted extensive experiments to better understand  the behavior of the proposed model.
\end{abstract}
\section{Introduction}\label{sec:intro}


Automatic lip-reading, also known as visual speech recognition, aims to recognize the speech content only based on visual information, especially the lip movements which are also named as visual speeches/sounds or visemes \cite{10.5555/2422356.2422395}. Lip-reading becomes a very challenging task for both human and machine, due to the ambiguity introduced by the one-to-many mapping \cite{Bear_2017} between viseme and phoneme. But a robust lip-reading system has a broad range of applications when the audio data is unavailable, such as silent speech control system \cite{10.1145/3242587.3242599}, assisting audio-based speech recognition in noisy environments \cite{8585066}, biometric authentication \cite{assael2016lipnet}.

Traditional approaches usually consist of a spatial feature extractor, such as discrete cosine transform 
 \cite{4041728,1230212,journals/ijst/PotamianosNISV01} to the lip Regions of Interest (RoIs), and followed by a sequential model which is usually a hidden markov model \cite{6047566,10.1016/j.specom.2007.11.002,865479}, to capture the temporal dynamics. More details about these older approaches are in \cite{ZHOU2014590}. Recently, two developments have been significantly improving the automatic lip-reading: the use of deep neural network models \cite{8461326,Weng2019_lipreading,9053841}, and the availability of a large scale dataset for training \cite{Chung16, yang2019lrw, Chung17}. Most deep-learning-based models usually consist of a frontend module and a backend module, which are similiar to the feature extractor and the sequential model in traditional approaches, respectively. However, using end-to-end training, the frontend module can extract more discriminative features than traditional extractor, and the backend module can capture more temporal information than the traditional approaches.

In this work, we focus on the word-level lip-reading and choose the Lip Reading in the Wild (LRW) \cite{Chung16} dataset as the benchmark. The current state-of-the-art (SOTA) performance on LRW is achieved by \cite{9053841}, which also is our baseline model. It consists of a 3D Convolutional layer followed by a 18-layer Residual Network (ResNet-18) \cite{7780459} and a Multi-Scale Temporal Convolutional Networks (MS-TCN), which are applied as the frontend and the backend, respectively. The final feature map at the sequence level is obtained by averaging the output of the backend along the time dimension. The process of merging feature maps in all time steps is called "\emph{Consensus}".

In this paper, we improve the performance of the current SOTA model. We revamp the frontend and the consensue to achieve a new SOTA. For the frontend, we proposed a novel  hierarchical pyramidal convolution (HPConv) to replace the standard 2D-convolution in the ResNet-18, which  is capable of processing the input with multiple spatial resolution. Moreover, our proposed model utilises the self-attention based consensus, which makes the frames related the annotated word play a greater role during classification. To the best of our knowledge, this is the first work introducing the multi-scale processing into the frontend and proposing changes in consensus for word-level lip-reading.


\section{Related Works}\label{sec:rel}

LRW is the first and the largest publicly available dataset with word-level label in English. It consists of short segments (1.16 seconds) from BBC news and talk shows. There are more than 1000 speakers and 500 target words, which is much higher than existing lip-reading databases used for word recognition. A total of 538766 segments in this dataset are split into 488766/25000/25000 for training/validation/testing usages. This is a quite challenging dataset due to the large number of  variations in head pose and illumination.
\begin{table}[htbp]
\begin{center}
\small
\setlength{\tabcolsep}{0.5pt}{
\begin{tabular}{c|c|c|c|c}
  \hline
  Method & Frontend & Backend & Consensus & Accuracy \\
  \hline
  \cite{Chung16} & VGG-M & - & - & 61.10 \\
  \cite{Chung17} & VGG-M & LSTM & Average & 76.20 \\
  \cite{Stafylakis2017} & 3D Conv+ResNet-34 & BLSTM & Average & 83 \\
  \cite{8461326} & 3D Conv+ResNet-34 & BGRU & Average & 83.4 \\
  \cite{STAFYLAKIS201822} & 3D Conv+ResNet-18 & BLSTM & Average & 84.3 \\
  \cite{DBLP:conf/bmvc/Wang19} & \makecell[c]{ResNet-34+\\3D DenseNet} & Conv-BLSTM & Average & 83.3 \\
  \cite{Weng2019_lipreading} & I3D*2 & BLSTM & Average & 84.07 \\
  \cite{xu2020discriminative} & \makecell[c]{3D Conv+\\P3D-ResNet-50} & BLSTM & Average & 84.48 \\
  \cite{9053841} &  3D Conv+ResNet-18 & MS-TCN & Average & 85.30 \\
 \cite{Liu2020} & \makecell[c]{3D Conv+ResNet-34\\+ST-GCN} & BGRU & Average & 84.25 \\
  \hline
\end{tabular}}
\caption{Review of  the existing models on LRW} \label{tab:summary_of_LRW}
\end{center}
\end{table}

Since LRW is released, numerous novel models were proposed for more powerful word-recognising abilities. We give a brief review of the previous models on LRW with their respective frontend type, backend type, consensus method and top-1 accuracy (abbreviated as accacy in the following), shown in Tabel \ref{tab:summary_of_LRW}.

\section{Our Approach}\label{sec:our}
\begin{figure}[htbp]
\centering
\includegraphics[width=0.96\linewidth]{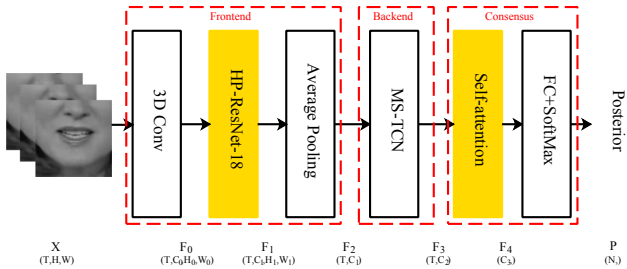}
\caption{The block-diagram of the proposed model. Our model can be divided to three main parts: the frontend module, the backend module and the consensus module. The input tensor and its shape of each module are given below. Our main contributions are highlighted in yellow.}
\label{fig:overview_of_our_approach}
\end{figure}
In this section, we present our proposed model in detail. As shown in Fig. \ref{fig:overview_of_our_approach}, our model can be divided to three main parts: the frontend module, the backend module and the consensus module. The frontend takes a grayscale sequence of lip RoIs $X\in \mathbb{R}^{\rm T\times H\times W}$ as input,  where $\rm T$ stands for the temporal dimension and $\rm H, W$ represent the height and width of the grayscale of lip respectively, and produce the feature $F_2\in \mathbb{R}^{\rm T\times C_1}$, which the spatial knowledge is summarised by applying the average pooling over the spatial dimensionality. After the frontend, the backend module is employed to model the temporal dynamics. The output $F_3\in \mathbb{R}^{\rm T\times C_2}$ is passed through the the consensus module to merge temporal information. Finally, the posterior probability of each word class $P$ is predicted by the ensuing a full connection layer and a SoftMax layer.

We maintain the multi-scale TCN in the baseline as the backend but change the frontend and the consensus. In the frontend, we replace the standrad convolution in the ResNet-18 with the hierarchical pyramidal convolution. Besides, the average based consensus is replaced by the self-attention based consensus.

\subsection{Hierarchical Pyramidal Convolution}\label{sec:hpconv}

\begin{figure}[htbp]
\centering
\begin{minipage}[b]{1.0\linewidth}
\centering
\includegraphics[width=0.76\linewidth]{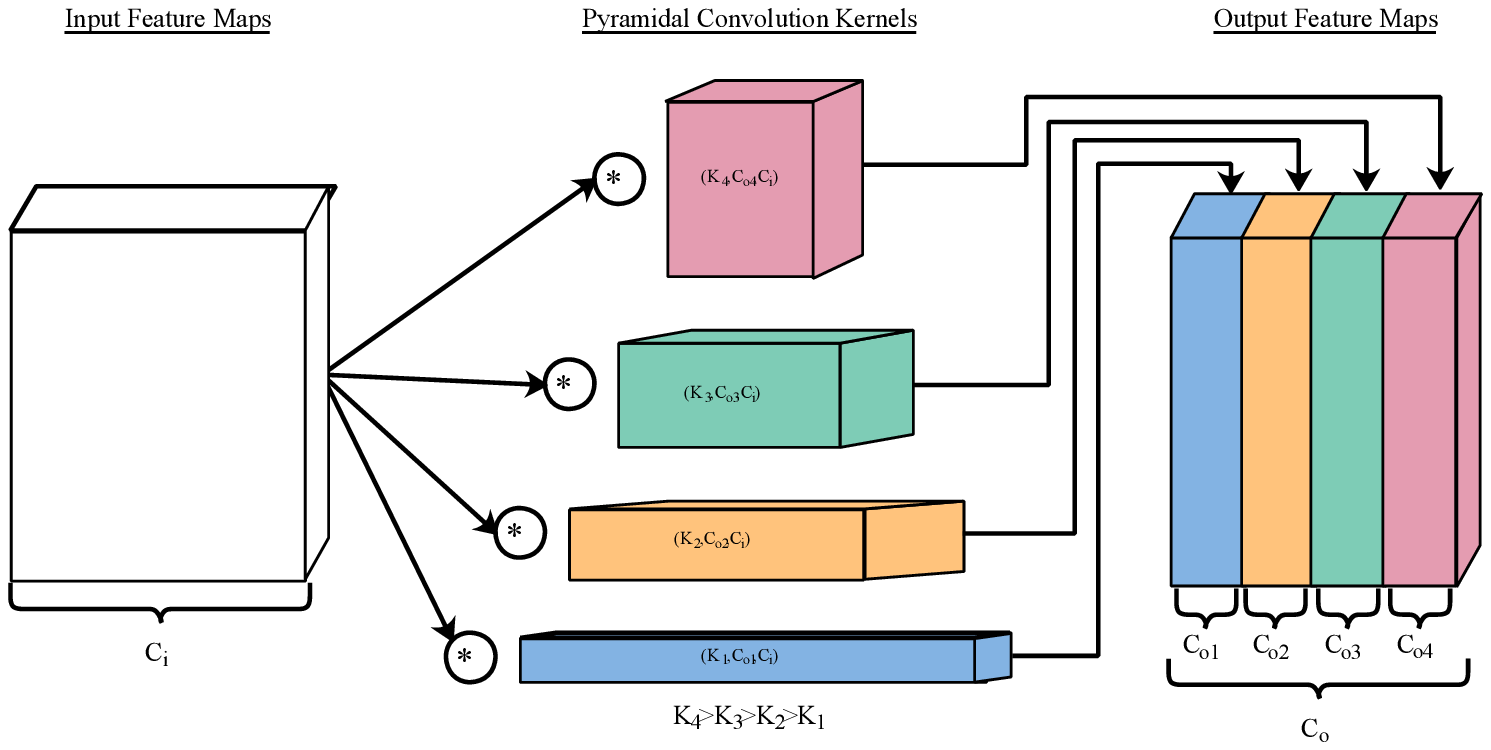}
\caption{Illustration of  the PyConv. The illustration is adapted from \cite{duta2020pyramidal}. Local and global feature maps are extracted from the input feature maps respectively. $\circledast$ denotes the convolution operation.}
\label{fig: pyconv}
\end{minipage}

\begin{minipage}[b]{1.0\linewidth}
\centering
\includegraphics[width=0.8\linewidth]{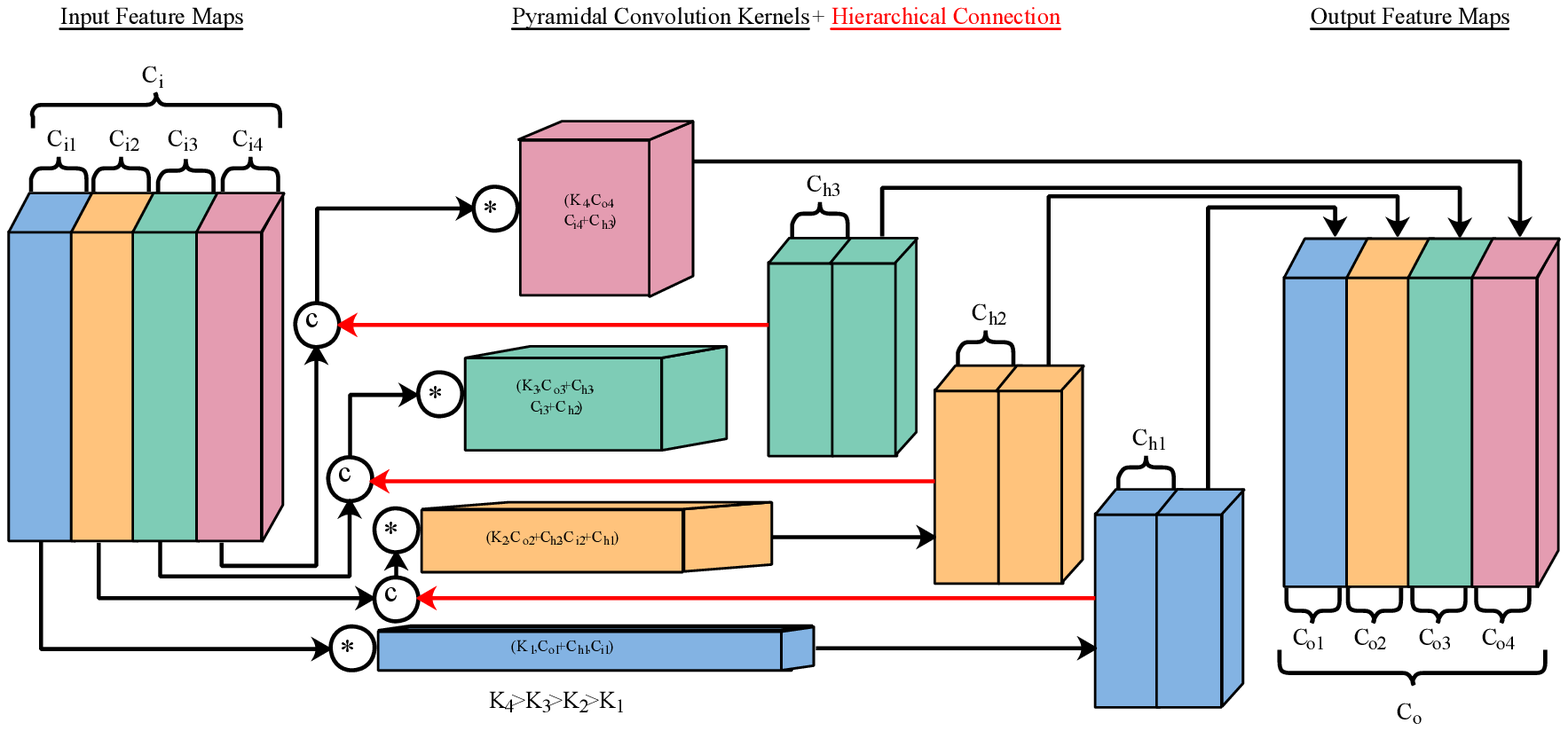}
\caption{Illustration of  the Proposed HPConv. Local feature maps also used to extract global feature maps. \copyright  denote the concatnation over channel dimension.}
\label{fig: hpconv}
\end{minipage}
\end{figure}

The ResNet-18 in the frontend of the baseline uses the standard 2D-convolution to extract spatial feature maps. The standard convolution contains a single type of kernel with a single spatial size $(K_1, K_1)$ (in the case of square kernels). All $C_o$ kernels have the same spatial resolution, which lead to a constant receptive field.

We analyze the error samples of the baseline, and find that the accuracy of word recognition increases with the number of viseme contained in the word. In other words, the model does not well on words with pool visemic content. This is reasonable, because the fewer visemes mean the fewer lip movements, which adds many extra difficulties for the model to classify sample correctly. Based on this, we proposed that applying different spatial size of kernels during the feature extraction can bring comlementary spatial context information, which enables the frontend can extract more distinct feature maps. These discriminative features help boosting the insight of the model into fine-grained lip movements and improve classification accuracy on words with few visemes.

To validate the effectiveness of the multi-sacle processing, we first introduce pyramidal convolution (PyConv) \cite{duta2020pyramidal} into the frontend. The PyConv, illustrated in Fig. \ref{fig: pyconv}, contains a pyramid with $n$  levels of different types of kernels (We set $n = 4$ as default in our experiments, which is consistent with the figure). The kernels at each level contains an increasing spatial sizes from the bottom of the pyramid to the top (We set $K_{1, 2, 3, 4} = 3,5,7,9$ as default in our experiments). The kernels with smaller spatial size can focus on details to extract feature maps with local context information, while the largger kernels can provides more global context information. The model can explore a best combination of different kernel types through learning. For every basic block of the ResNet-18, we replace the second standard convolution layer to the PyConv. We call this modification as the Pyramidal ResNet-18 (Py-ResNet-18).

Based on the PyConv, we propose hierarchical pyramidal convolution (HPConv), illustrated in Fig. \ref{fig: hpconv}. The most innovative point is that we establish a hierarchical connection between adjacent layers of the pyramid (Red lines in \ref{fig: hpconv}). As mentioned above, the local and global feature maps in the PyConv are extracted from the input feature maps respectively. And with the hierarchical connection, the local feature map is used as a part of the output, also an input for the global feature extracttion. We proposed that this bottom-up information aggregation can further improve the classification performation of the model, specially on words with few visemes. For every basic block of the ResNet-18, we replace the second standard convolution layer to the HPConv. We call this modification as the Hierarchical Pyramidal ResNet18 (HP-ResNet-18).

\subsection{Self-attention based consensus}\label{sec:att}
\begin{figure}[htbp]
\centering
\includegraphics[width=0.96\linewidth]{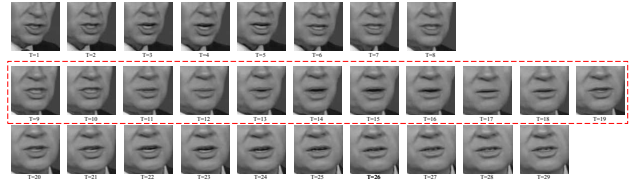}
\caption{An example of a video sample annotated as "ABOUT". Only frames at time step $T = 9\sim 19$ are related with the word "ABOUT".}
\label{fig:video_sample_of_ABOUT}
\end{figure}

 The most popular consensus method currently is to average over all the time steps, as shown in Tabel \ref{tab:summary_of_LRW}. For the average based consensus, given the feature maps at the frame level $F_3\in \mathbb{R}^{T\times C_2}$,  the final feature at the sequence level $F_4\in\mathbb{R}^{C_3}$ is calculated as follows:
\begin{align}
F_4 = \frac{\sum_{t=0}^{T-1}F_{3,t}}{T}
\end{align}

The average based consensus assumes that every frame provides an equal contributoin to the final decision, which is not the case in practice. As shown in Fig. \ref{fig:video_sample_of_ABOUT}, the video sample annotated as "ABOUT" includes $29$ frames in total, but only frames at time step $T = 9\sim 19$ are related with the word "ABOUT", shown in the red boxes. In practice, the boundaries of individual words are hard to get too. Based on this, We popose a self-attention \cite{vaswani2017attention} based consensus method to ensure the model pay more attention to the frames which are related with annotated word, but less to other irrelevant frames. Our self-attention based consensus can be expressed as:
\begin{align}
& Q_n,K_n,V_n  = F_3W^Q_n,F_3W^K_n,F_3W^V_n\\
& H_n = A_n^{\mathsf{T}}V_n={\rm SoftMax}(\frac{\sum_{t=0}^{T-1}Q_{n, {t}}K_n^{\mathsf{T}}}{{T}\sqrt{d_k}})V_n\\
& F_4 = W^O{\rm Concat}(H_0,\cdots,H_{N-1})+\frac{\sum_{t=0}^{T-1}F_{3,{t}}}{T}
\end{align}
where the projection matrices $W^Q_n\in \mathbb{R}^{C_2\times d_k}$, $W^K_n\in \mathbb{R}^{C_2\times d_k}$, $W^V_n\in \mathbb{R}^{C_2\times d_v}$ and $W^O\in \mathbb{R}^{Nd_v\times C_3}$, the attention weight $A_n\in \mathbb{R}^{T}$. In this work we employ $N=8$ heads and $ d_k=d_v=64$, same with \cite{vaswani2017attention}. Even though each head has a different focus, the attention weight on all irrelevant frames should be $0$, which helps the model to ignore noisy information for better classification performation. 


\section{Experiment Results}\label{sec:exp_res}

\begin{table}[htbp]
\begin{center}
\small
\setlength{\tabcolsep}{3pt}{
\begin{tabular}{c|c|c|c|c}
  \hline
  Model & Frontend & Consensus & Boundary &Accuracy \\
  \hline
  baseline & ResNet-18& Average & F & 85.3 \\
  N1 & ResNet-18  & Self-attention & F & 86.47\\
  N2 & Py-ResNet-18 & Average & F & 85.88\\
  N3 & HP-ResNet-18 & Average & F & 86.45\\
  \textbf{N4 (Our)} & HP-ResNet-18 & Self-attention & F & \textbf{86.83}\\
  \hline
  N5 & ResNet-18  & Average & T & 88.60\\
  N6 & ResNet-18  & Self-attention & T & 88.59\\
  \hline
\end{tabular}}
\caption{A comparison of the performance between the baseline and our models. Our model attain the state-of-the-art on LRW. 3D Conv in the frontend is omitted for simplicity.} \label{tab:comparison_all}
\end{center}
\end{table}

\begin{figure*}[htbp]
\begin{minipage}[b]{0.38\linewidth}
  \centering
  \centerline{\includegraphics[width=0.96\linewidth]{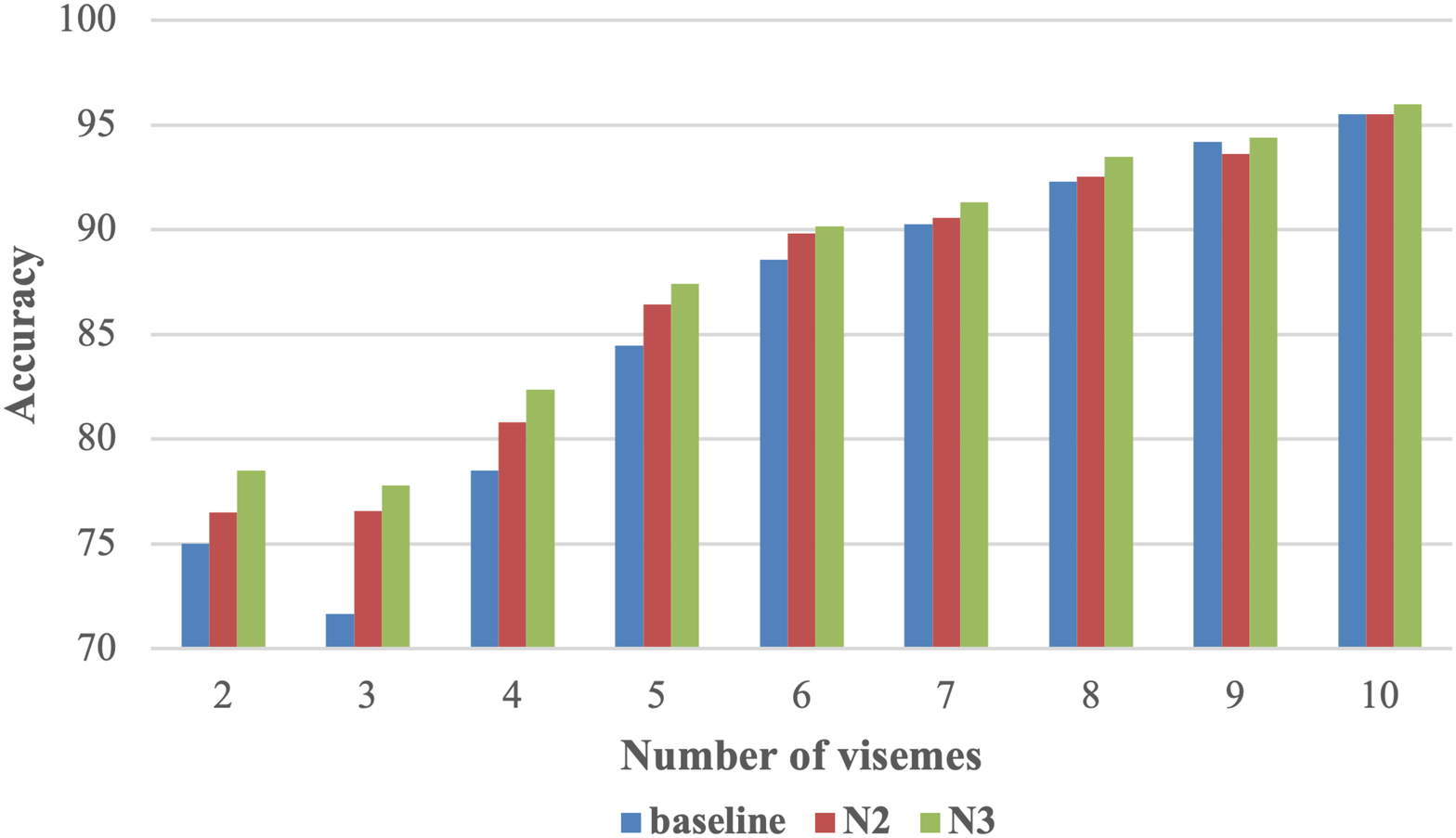}}
  \caption{A comparison of  the baseline and two our models (N2 and N3) on the 9 categories.}
\label{fig: acc_vs_viseme}
\end{minipage}
\begin{minipage}[b]{0.3\linewidth}
  \centering
  \centerline{\includegraphics[width=0.96\linewidth]{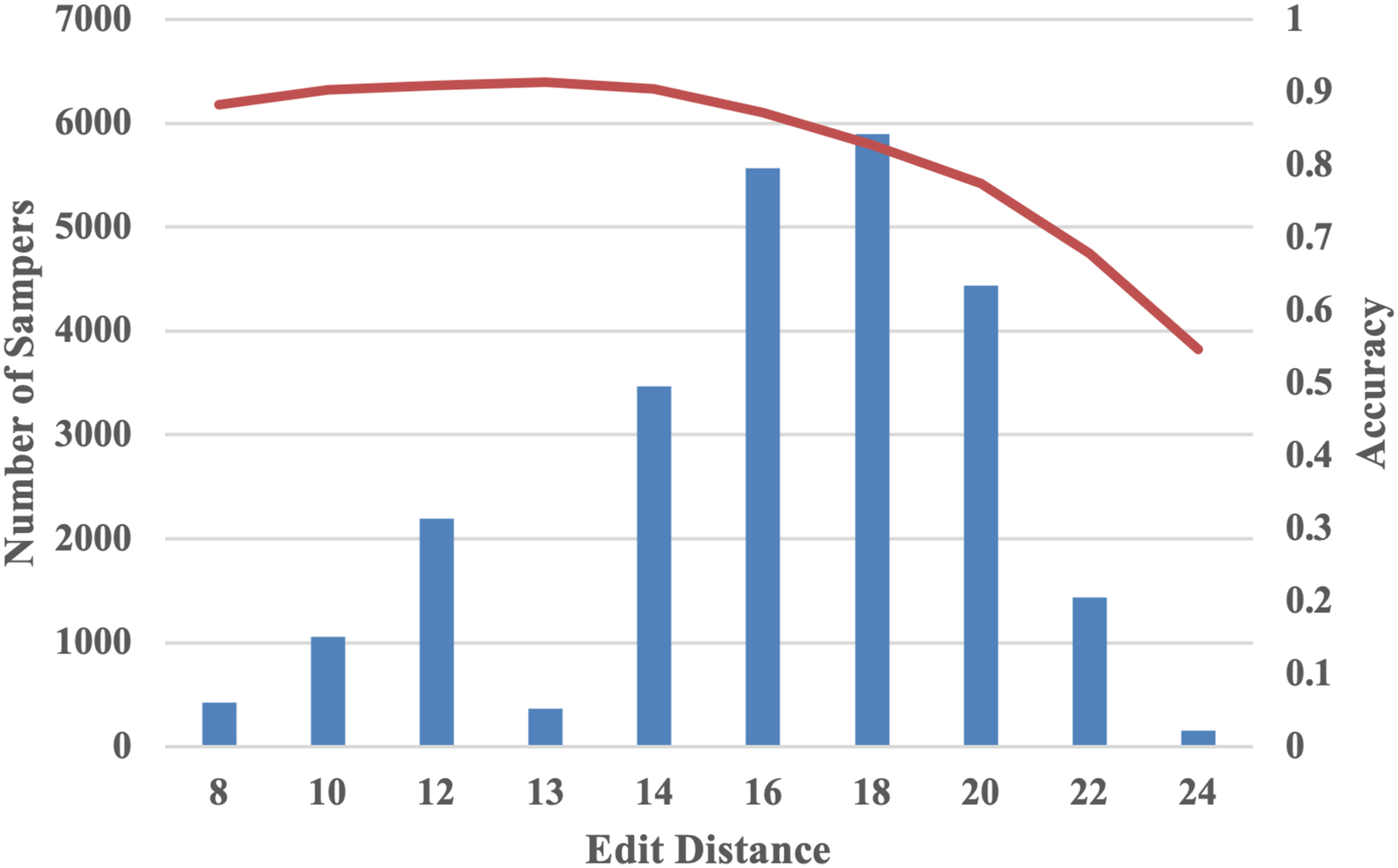}}
  \caption{Statistical results of the baseline on different edit distances.}
\label{fig: dis_vs_num_acc_average}
\end{minipage}
\begin{minipage}[b]{0.3\linewidth}
  \centering
  \centerline{\includegraphics[width=0.96\linewidth]{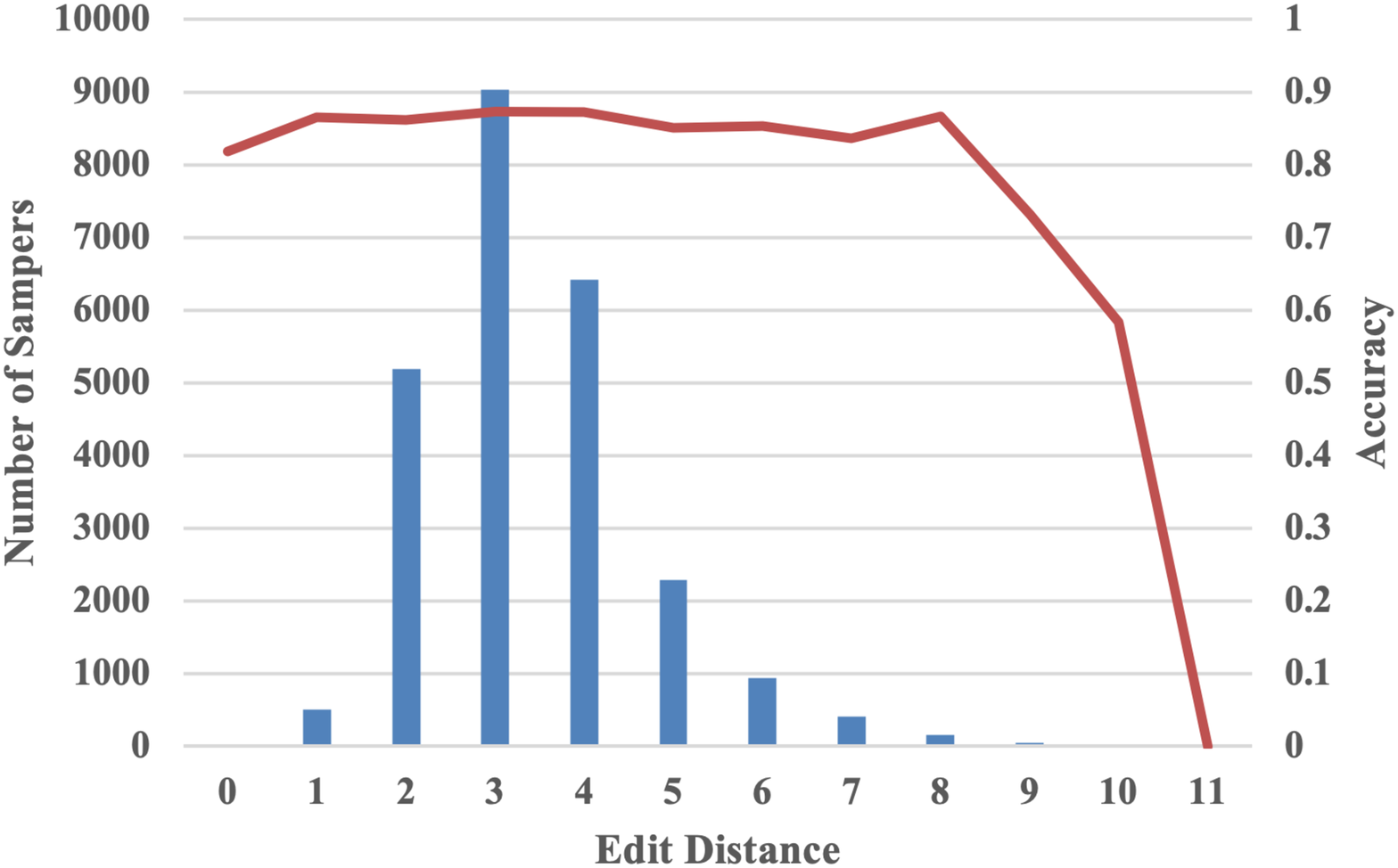}}
  \caption{Statistical results of N1 on different edit distances.}
\label{fig: dis_vs_num_acc_attention}
\end{minipage}
\end{figure*}

In this section, we compare our model with the baseline model. And to understand the contribution of different parts of our model better, we also analyze the results of the ablation experiment. We pre-process each video process and train all models following the same method as the baseline. The readers are referred to \cite{9053841} for more details.

%
%
%


Table \ref{tab:comparison_all} lists the results of all models. Compared to the baseline model, our proposed model (denoted by N4) achieves an absolute improvement of $1.53\%$ in classification accuracy, which means our model attain the state-of-the-art by a wide margin on LRW.

\subsection{Discussion about Proposed HPConv}\label{sec:dis_hpconv}

To verify the effectiveness of our proposed HPConv on classification performance, we present the result of the model using only the HP-ResNet-18 in Tabel \ref{tab:comparison_all} (denoted by N3), along with the result of the model using only the Py-ResNet-18 (denoted by N2), as it constitutes the starting point for our work on the multi-scale process in lip-reading. Compared with N2, N3 and the baseline, applying the multi-scale kernel significantly enhances the classification performance of the model in the spatial feature extraction. In other words, our proposed HPConv benifits the model much more than the PyConv.

To further explore why our HPConv can outperform the baseline model and the PyConv, we analyze the error samples of different models. Based on the number of viseme in the annotated word, we divide the whole test set into 9 categories and present the classification accuracies of different models on these 9 categories in Fig.\ref{fig: acc_vs_viseme}. We come to a conclusion that the PyConv extract spactial features in different resolution, which enhances the ability of model to capture imperceptible lip movements. As a result, the classification accuracy of the model on words with few visemes is significantly improved. What's more, our proposed HPConv introduces the hierarchical connection from local to global information, which further improves classification accuracy on words with few visemes.

\subsection{Discussion about Proposed Self-attention Based Consensus}\label{sec:dis_att}

One of the most significant differences between our method and previous methods is the proposed self-attention based consensus. It ensures that the model pays more attention on the relevant frames durning classification. To verify its effectiveness, we present the result of the model using only the self-attention based consensus in Table \ref{tab:comparison_all} (deneted by N1). Compared with N1 and the baseline, we can say that the self-attention based consensus improves the classification performance.

To further explore why our self-attention based consensus can outperform the average based consensus, we retrain the baseline and N1 using word boundary offered by \cite{Chung16}. The results are denoted as N5 and N6, respectively. The major difference is only on applying average based consensus or self-attention based consensus on the the frames which are related the annotated word. By comparing N5 with N6, we find that, in the situation of manual word boundary is used, self-attention based consensus basically does not work. This may result from that, the attention weight is a learned "soft word boundary". It does the similar thing as manual word boundary, but not as accurate.

To verify our assumption, we categorize the all test samples by the edit distance of the manual word boundary and the learned word boundary, and count the number of samples and classification accuracy of each category. For the average based consensus, the learned word boundary $B_{avg}=[1,\cdots, 1]^{\mathsf{T}} \in  \mathbb{R}^{T}$. For the self-attention based consensus, the learned word  $B_{att}=\mathrm{u}(\sum_{n=0}^{N-1}H_n/N-\alpha)$, where $\mathrm{u}(\cdot)$ is the unit step function and the threshold constant $\alpha=0.01$.

The statistical results of the baseline and N1 are shown in Fig. \ref{fig: dis_vs_num_acc_average} and Fig. \ref{fig: dis_vs_num_acc_attention}. From these two figures, we conclude that accuracy is dropping with the increasing of the edit distance and the self-attention based consensus can learn a more precise word boundary. 

\section{Conclusion}\label{sec: con}

In this work, we have presented the HPConv and the self-attention based consensus, replacing the standrad convolution and the average based consensus most commonly used in lip-reading models, respectively. Extensive experiments and analysis empirically validate that our proposed HPConv improve the model's perception of slight lip movements and the self-attention based consensus ensure the model pay more attention to the relevant frames. Combining them results in a new SOTA performance.


\bibliographystyle{IEEEbib}
\bibliography{main}

\begin{thebibliography}{10}

\bibitem{10.5555/2422356.2422395}
Sarah~L. Taylor, Moshe Mahler, Barry-John Theobald, and Iain Matthews,
\newblock ``Dynamic units of visual speech,''
\newblock in {\em Proceedings of the ACM SIGGRAPH/Eurographics Symposium on
  Computer Animation}, Goslar, DEU, 2012, SCA '12, p. 275–284, Eurographics
  Association.

\bibitem{Bear_2017}
Helen~L Bear and Richard Harvey,
\newblock ``Phoneme-to-viseme mappings: the good, the bad, and the ugly,''
\newblock {\em Speech Communication}, vol. 95, pp. 40–67, Dec 2017.

\bibitem{10.1145/3242587.3242599}
Ke~Sun, Chun Yu, Weinan Shi, Lan Liu, and Yuanchun Shi,
\newblock ``Lip-interact: Improving mobile device interaction with silent
  speech commands,''
\newblock in {\em Proceedings of the 31st Annual ACM Symposium on User
  Interface Software and Technology}, New York, NY, USA, 2018, UIST '18, p.
  581–593, Association for Computing Machinery.

\bibitem{8585066}
T.~{Afouras}, J.~S. {Chung}, A.~{Senior}, O.~{Vinyals}, and A.~{Zisserman},
\newblock ``Deep audio-visual speech recognition,''
\newblock {\em IEEE Transactions on Pattern Analysis and Machine Intelligence},
  pp. 1--1, 2018.

\bibitem{assael2016lipnet}
Yannis~M. Assael, Brendan Shillingford, Shimon Whiteson, and Nando de~Freitas,
\newblock ``Lipnet: End-to-end sentence-level lipreading,'' 2016.

\bibitem{4041728}
X.~{Hong}, H.~{Yao}, Y.~{Wan}, and R.~{Chen},
\newblock ``A pca based visual dct feature extraction method for lip-reading,''
\newblock in {\em 2006 International Conference on Intelligent Information
  Hiding and Multimedia}, 2006, pp. 321--326.

\bibitem{1230212}
G.~{Potamianos}, C.~{Neti}, G.~{Gravier}, A.~{Garg}, and A.~W. {Senior},
\newblock ``Recent advances in the automatic recognition of audiovisual
  speech,''
\newblock {\em Proceedings of the IEEE}, vol. 91, no. 9, pp. 1306--1326, 2003.

\bibitem{journals/ijst/PotamianosNISV01}
Gerasimos Potamianos, Chalapathy Neti, Giridharan Iyengar, Andrew~W. Senior,
  and Ashish Verma,
\newblock ``A cascade visual front end for speaker independent automatic
  speechreading.,''
\newblock {\em I. J. Speech Technology}, vol. 4, no. 3-4, pp. 193--208, 2001.

\bibitem{6047566}
V.~{Estellers}, M.~{Gurban}, and J.~{Thiran},
\newblock ``On dynamic stream weighting for audio-visual speech recognition,''
\newblock {\em IEEE Transactions on Audio, Speech, and Language Processing},
  vol. 20, no. 4, pp. 1145--1157, 2012.

\bibitem{10.1016/j.specom.2007.11.002}
Xu~Shao and Jon Barker,
\newblock ``Stream weight estimation for multistream audio-visual speech
  recognition in a multispeaker environment,''
\newblock {\em Speech Commun.}, vol. 50, no. 4, pp. 337–353, Apr. 2008.

\bibitem{865479}
S.~{Dupont} and J.~{Luettin},
\newblock ``Audio-visual speech modeling for continuous speech recognition,''
\newblock {\em IEEE Transactions on Multimedia}, vol. 2, no. 3, pp. 141--151,
  2000.

\bibitem{ZHOU2014590}
Ziheng Zhou, Guoying Zhao, Xiaopeng Hong, and Matti Pietikäinen,
\newblock ``A review of recent advances in visual speech decoding,''
\newblock {\em Image and Vision Computing}, vol. 32, no. 9, pp. 590 -- 605,
  2014.

\bibitem{8461326}
S.~{Petridis}, T.~{Stafylakis}, P.~{Ma}, F.~{Cai}, G.~{Tzimiropoulos}, and
  M.~{Pantic},
\newblock ``End-to-end audiovisual speech recognition,''
\newblock in {\em 2018 IEEE International Conference on Acoustics, Speech and
  Signal Processing}, 2018, pp. 6548--6552.

\bibitem{Weng2019_lipreading}
Xinshuo Weng and Kris Kitani,
\newblock ``{Learning Spatio-Temporal Features with Two-Stream Deep 3D CNNs for
  Lipreading},''
\newblock {\em BMVC}, 2019.

\bibitem{9053841}
B.~{Martinez}, P.~{Ma}, S.~{Petridis}, and M.~{Pantic},
\newblock ``Lipreading using temporal convolutional networks,''
\newblock in {\em 2020 IEEE International Conference on Acoustics, Speech and
  Signal Processing}, 2020, pp. 6319--6323.

\bibitem{Chung16}
J.~S. Chung and A.~Zisserman,
\newblock ``Lip reading in the wild,''
\newblock in {\em Asian Conference on Computer Vision}, 2016.

\bibitem{yang2019lrw}
Dalu Feng Mingmin Yang Chenhao Wang Jingyun Xiao Keyu Long Shiguang Shan
  Xilin~Chen Shuang~Yang, Yuanhang~Zhang,
\newblock ``Lrw-1000: A naturally-distributed large-scale benchmark for lip
  reading in the wild,''
\newblock in {\em 2019 14th IEEE International Conference on Automatic Face \&
  Gesture Recognition}. IEEE, 2019, pp. 1--8.

\bibitem{Chung17}
J.~S. Chung, A.~Senior, O.~Vinyals, and A.~Zisserman,
\newblock ``Lip reading sentences in the wild,''
\newblock in {\em IEEE Conference on Computer Vision and Pattern Recognition},
  2017.

\bibitem{7780459}
K.~{He}, X.~{Zhang}, S.~{Ren}, and J.~{Sun},
\newblock ``Deep residual learning for image recognition,''
\newblock in {\em 2016 IEEE Conference on Computer Vision and Pattern
  Recognition}, 2016, pp. 770--778.

\bibitem{Stafylakis2017}
Themos Stafylakis and Georgios Tzimiropoulos,
\newblock ``Combining residual networks with lstms for lipreading,''
\newblock in {\em Proc. Interspeech 2017}, 2017, pp. 3652--3656.

\bibitem{STAFYLAKIS201822}
Themos Stafylakis, Muhammad~Haris Khan, and Georgios Tzimiropoulos,
\newblock ``Pushing the boundaries of audiovisual word recognition using
  residual networks and lstms,''
\newblock {\em Computer Vision and Image Understanding}, vol. 176-177, pp. 22
  -- 32, 2018.

\bibitem{DBLP:conf/bmvc/Wang19}
Chenhao Wang,
\newblock ``Multi-grained spatio-temporal modeling for lip-reading,''
\newblock in {\em 30th British Machine Vision Conference}, 2019, p. 276.

\bibitem{xu2020discriminative}
Bo~Xu, Cheng Lu, Yandong Guo, and Jacob Wang,
\newblock ``Discriminative multi-modality speech recognition,''
\newblock in {\em Proceedings of the IEEE/CVF Conference on Computer Vision and
  Pattern Recognition}, 2020, pp. 14433--14442.

\bibitem{Liu2020}
Hong Liu, Zhan Chen, and Bing Yang,
\newblock ``{Lip Graph Assisted Audio-Visual Speech Recognition Using
  Bidirectional Synchronous Fusion},''
\newblock in {\em Proc. Interspeech 2020}, 2020, pp. 3520--3524.

\bibitem{duta2020pyramidal}
Ionut~Cosmin Duta, Li~Liu, Fan Zhu, and Ling Shao,
\newblock ``Pyramidal convolution: Rethinking convolutional neural networks for
  visual recognition,''
\newblock {\em arXiv preprint arXiv:2006.11538}, 2020.

\bibitem{vaswani2017attention}
Ashish Vaswani, Noam Shazeer, Niki Parmar, Jakob Uszkoreit, Llion Jones,
  Aidan~N Gomez, {\L}ukasz Kaiser, and Illia Polosukhin,
\newblock ``Attention is all you need,''
\newblock {\em Advances in neural information processing systems}, vol. 30, pp.
  5998--6008, 2017.

\end{thebibliography}

\end{document}